%% file: main.tex
\documentclass[10pt,twocolumn,letterpaper]{article}

\usepackage{cvpr}              


\definecolor{cvprblue}{rgb}{0.21,0.49,0.74}
\usepackage[pagebackref,breaklinks,colorlinks,allcolors=cvprblue]{hyperref}
\usepackage{multirow}
\usepackage{diagbox}
\usepackage{adjustbox}

\def\paperID{2} 
\def\confName{CVPR}
\def\confYear{2025}

\title{On the Suitability of Reinforcement Fine-Tuning to Visual Tasks}

\author{
Xiaxu Chen\textsuperscript{1,2}\thanks{Email: \texttt{13373873609@163.com}} \quad
Wei Li\textsuperscript{1}\thanks{Email: \texttt{liwei1@sensetime.com}} \quad
Chunxu Liu\textsuperscript{1,3} \quad
Chi Xie\textsuperscript{1,4} \\
Xiaoyan Hu\textsuperscript{1} \quad
Chengqian Ma\textsuperscript{1} \quad
Feng Zhu\textsuperscript{1} \quad
Rui Zhao\textsuperscript{1}\thanks{Corresponding author. Email: \texttt{zhaorui@sensetime.com}}\\
\textsuperscript{1}SenseTime Research \quad
\textsuperscript{2}Beijing Institute of Technology \quad
\textsuperscript{3}Nanjing University \quad
\textsuperscript{4}Tongji University \\
}

\begin{document}
\maketitle
\input{sec/0_abstract}    
\input{sec/1_intro}
\input{sec/2_related}
\input{sec/3_method}

\input{sec/4_experiment}

\input{sec/5_conclusion}

{
    \small
    \bibliographystyle{ieeenat_fullname}
    \bibliography{main}
}


\end{document}

%% file: sec/0_abstract.tex
\begin{abstract}
Reinforcement Fine-Tuning (RFT) is proved to be greatly valuable for enhancing the reasoning ability of LLMs.
Researchers have been starting to apply RFT to MLLMs, hoping it will also enhance the capabilities of visual understanding.
However, these works are at a very early stage and have not examined how suitable RFT actually is for visual tasks.
In this work, we endeavor to understand the suitabilities and limitations of RFT for visual tasks, through experimental analysis and observations.
We start by quantitative comparisons on various tasks, which shows RFT is generally better than SFT on visual tasks. 
To check whether such advantages are brought up by the reasoning process, we design a new reward that encourages the model to ``think'' more, whose results show more thinking can be beneficial for complicated tasks but harmful for simple tasks.
We hope this study can provide more insight for the rapid advancements on this topic.
\end{abstract}

%% file: sec/1_intro.tex
\section{Introduction}
\label{sec:intro}
Recently, Reinforcement Fine-Tuning (RFT) has demonstrated remarkable effectiveness on Large Language Models (LLMs) such as DeepSeek-R1~\cite{guo2025deepseekr1}. By incentivizing the model to engage in more extensive ``thinking'' during training and inference, RFT significantly enhances its reasoning capabilities for addressing complex language tasks.
Relevant techniques include Reinforcement Learning with Human Feedbacks (RLHF) and Reinforcement Learning with Verifiable Rewards (RLVR), which utilizes human preferences or objectively verifiable outcomes as rewards for reinforcement learning.

A natural question emerges: can RFT similarly augment Multimodal Large Language Models (MLLMs), particularly in the realm of visual reasoning? Recent studies~\cite{huang2025visionr1,liu2025visualrft,meng2025mmeureka,zhou2025r1visualthinker} have investigated the application of RFT to MLLMs, achieving superior performance on tasks that explicitly demand robust reasoning skills. These efforts have underscored RFT's strengths in Few-Shot Classification, Object Detection, and Reasoning Grounding, surpassing the capabilities of Supervised Fine-Tuning (SFT). Nevertheless, the extent of RFT's applicability to visual tasks remains largely unexplored.

In this study, we examine the impact of RFT on MLLMs, contrasting it with prior approaches such as MLLMs trained with SFT. We begin by implementing RFT on MLLMs and evaluating their performance against SFT across various computer vision tasks from perception classification tasks to those need visual reasoning. Notably, RFT consistently delivers substantial improvements on specific tasks, often outperforming SFT by a wide margin. 


We then explore whether the performance advantage of RFT over SFT stems from improved reasoning. To investigate this, we introduce a \textbf{Normalized Length Reward} in the RFT framework, encouraging the model to produce lengthier intermediate outputs and engage in prolonged ``thinking''. This adjustment enhance the performance on complicated tasks requiring explicit reasoning but decrease it on perception classification tasks, suggesting that the gains are partially attributable to enhancing model’s structured reasoning capabilities from RFT. Besides, disabling the thinking process during inference consistently impairs MLLM performance. We therefore conclude that \textit{current computer vision tasks demands different degrees of reasoning according to their task nature}, and insights gained from RFT on LLMs cannot be directly applied to visual domains.

Our contributions are summarized as follows:

\begin{enumerate}
\item We demonstrate that Reinforcement Fine-Tuning (RFT) outperforms SFT across a range of computer vision tasks, from basic classification to those requiring visual reasoning.


\item By encouraging MLLMs to think longer using the Normalized Length Reward in RFT, MLLMs obtain reasonable thinking process and stronger performance on some complicated tasks that require explicit reasoning.

\item We find that encouraging longer thinking process can be harmful on some simple visual tasks, which shows RFT from LLMs require more improvements before applied to visual tasks.

\end{enumerate}

%% file: sec/2_related.tex
\section{Related Works}
\label{sec:related}

\noindent \textbf{Multimodal Large Language Models.}
Multimodal Large Language Models (MLLMs)~\cite{hurst2024gpt, liu2023llava, bai2025qwen25vl, chen2024internvl} integrate visual encoders with Large Language Models (LLMs) to enable visual perception and achieve remarkable performance on multimodal tasks~\cite{liu2024mmbench, li2024seed, fu2023mme}. A typical MLLM consists of a vision encoder, an LLM, and a visual projector that maps visual tokens into the semantic space of LLM. Leveraging this architecture, MLLMs have been applied to a wide range of vision and language tasks, including image classification~\cite{ouali2024discriminative, mitra2024sparse}, object segmentation~\cite{lai2024lisa}, object detection~\cite{pi2023detgpt}, information retrieval~\cite{liu2024lamra}, and visual question answering~\cite{hurst2024gpt, liu2023llava, bai2025qwen25vl, chen2024internvl}. 
In this paper, we further explore the versatility of Reinforcement Fine-Tuning (RFT) in enhancing the test-time scaling ability of MLLMs on various computer vision tasks.

\noindent \textbf{Reasoning in MLLMs.}
Since the recent surge of reasoning in Large Language Models like Openai-o1 and DeepSeek-R1~\cite{guo2025deepseekr1}, the community has been trying to achieve a similar reasoning process for MLLMs. They utilize Reinforcement Learning with Verifiable Rewards, a training approach to enhance language models in tasks with objectively verifiable outcomes.
Exploration in this area is still at a very early stage and remains highly immature.
Among them, R1-V explores how to transplant R1 directly to MLLMs.
VisualThinker-R1-Zero~\cite{zhou2025r1visualthinker} claims to be the first to produce ``aha moment'' and increased response length for MLLMs, by performing RFT on base models without instruction tuning.
Visual-RFT~\cite{liu2025visualrft} finds that reinforcement fine-tuning is more powerful than supervised fine-tuning on a wide range of tasks like few-shot classification and detection.
Such concurrent works all try to transplant R1 from LLMs to MLLMs and prove how powerful R1 is. Differently, this work tries to examine where RFT is suitable and where not on traditional computer vision tasks.

%% file: sec/3_method.tex
\section{Method}
\label{sec:method}

\begin{figure*}[t]
    \centering
    \includegraphics[width=0.9\linewidth]{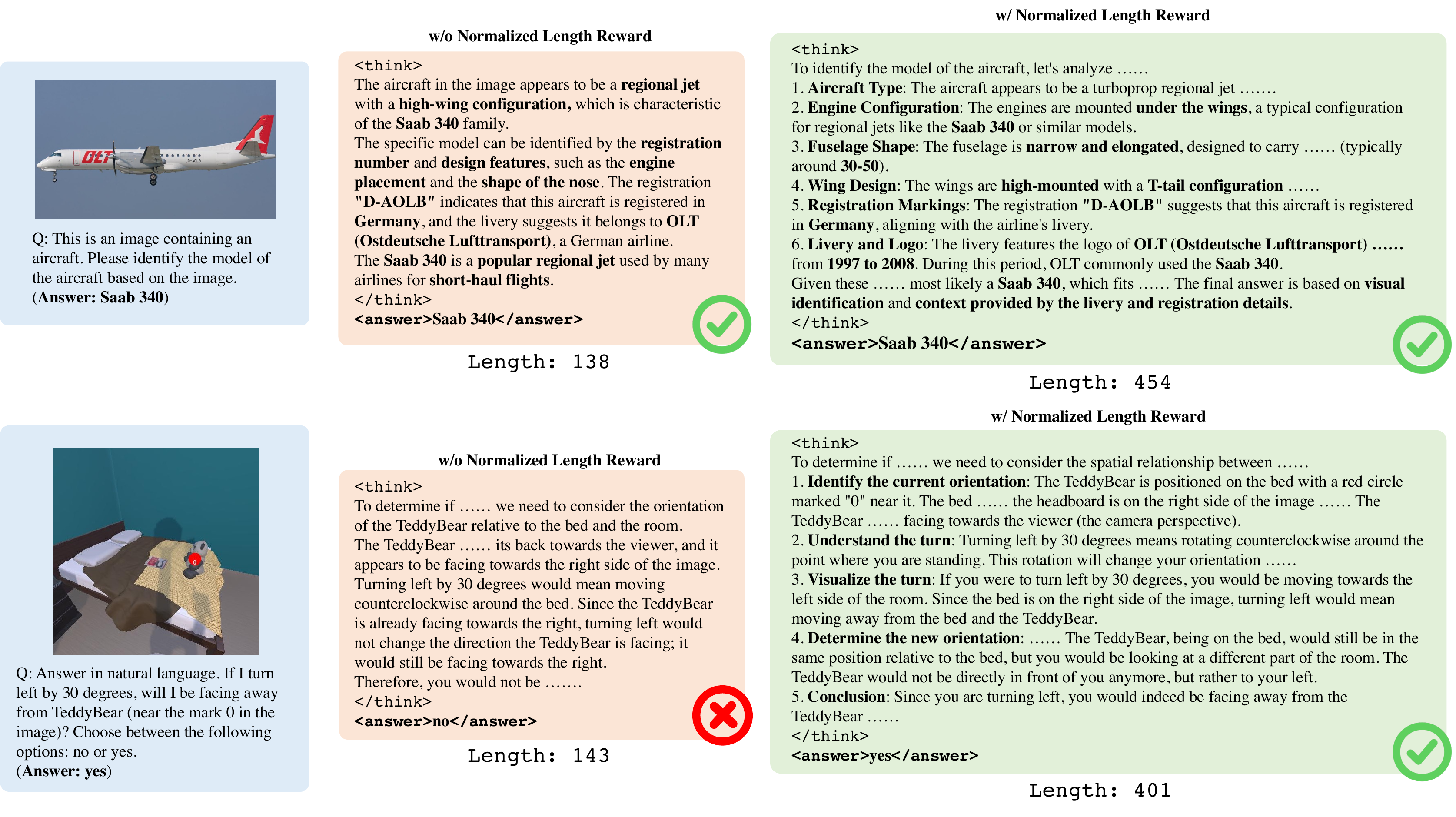}
    \caption{\textbf{Qualitative Comparison Between w/o and w/ Normalized Length Reward.} \textit{Best view in zoom.}}
    \label{fig:vis}
\end{figure*}

\noindent \textbf{Reinforcement Learning in Large Models.}
Reinforcement Learning with Verifiable Rewards \cite{shao2024deepseekmath,guo2025deepseekr1,team2025kimik15} is a training paradigm aimed at improving language models in domains where correctness can be objectively verified, such as mathematics and programming. In contrast to Reinforcement Learning from Human Feedback (RLHF) \cite{ouyang2022rlhf}, which depends on a learned reward model, RLVR evaluates outputs using a direct verification function. This eliminates the need for an intermediary reward model while ensuring that the optimization process remains closely tied to the intrinsic correctness measures of tasks. Given an input question $q$, the policy model $\pi_{\theta}$ generates a response $o$ and receives a verifiable reward accordingly. The training objective optimized in RLVR is expressed as follows:
\begin{align}
    & \max_{\pi_{\theta}} \mathbb{E}_{o \sim \pi_{\theta}(q)} \left[ R_{\text{RLVR}}(q, o) \right] \\
    & = \left[ R(q, o) - \beta \text{KL}[\pi_{\theta}(o|q) \parallel \pi_{\text{ref}}(o|q)] \right],
\end{align}
where $\pi_{\text{ref}}$ represents the pre-optimization reference model, $R$ denotes the reward function used for verification, and $\beta$ is a hyperparameter that regulates the KL-divergence. The reward function $R$ evaluates a given question-output pair $(q,o)$ by determining whether the predicted response $o$ aligns with the ground-truth answer.

\noindent \textbf{Group Relative Policy Optimization.}
DeepSeek R1-Zero~\cite{guo2025deepseekr1} removes the need for supervised fine-tuning by adopting reinforcement learning for post-training. Specifically, it utilizes Group Relative Policy Optimization (GRPO)~\cite{shao2024deepseekmath}, which differs from conventional reinforcement learning methods like PPO~\cite{schulman2017ppo} that rely on a critic model for policy evaluation. Instead of using a separate critic, GRPO directly assesses and ranks multiple candidate responses. Given an input query $q$, the model generates $G$ potential answers ${o_1, o_2, ..., o_G}$ based on its current policy $\pi_{\theta_{\text{old}}}$. The rewards associated with these responses, denoted as ${r_1, r_2, ..., r_G}$, are then computed. GRPO normalizes these rewards by calculating their mean and standard deviation:
\begin{align}
A_i &= \frac{r_i - \text{mean}(\{r_1, \dots, r_G\})}{\text{std}(\{r_1, \dots, r_G\})},
\end{align}
where $A_i$ indicates the relative quality of the $i$-th response. This approach guides the model toward generating higher-quality responses by prioritizing those with relatively superior rewards.

\noindent \textbf{Verifiable Rewards for Vision Tasks.}
The reward model is a key step in reinforcement learning (RL) that aligns models with preference alignment algorithms, which can be as straightforward as a verification function that checks for exact matches between predictions and ground-truth answers.
The RL training process in the recent DeepSeek-R1~\cite{guo2025deepseekr1} model achieves a significant improvement in the model's reasoning ability through the verifiable reward design.
To transfer this strategy to the visual domain, we design different rule-based verifiable reward functions for various visual perception tasks.

\noindent \textbf{Normalized Length Reward.}
In VisualThinker~\cite{zhou2025r1visualthinker}, researchers observed that using a fixed-length reward of $+0.001$ per additional generated token led to an increase in response length without improving accuracy. The model exploited the reward by generating repetitive content rather than enhancing its reasoning process. To address this issue, w e apply sigmoid normalization to the length reward, stabilizing its value between 0 and 1, as follows:
\begin{align}
    R = \frac{1}{1+\lambda \cdot \exp(-(L-L_0))},
\end{align}
where $\lambda$ is a scaling factor set to 1 in our experiments, and $L_0$ is a predefined length fixed at $100$ across all experiments. Therefore, the length reward is kept within a reasonable range, as the diminishing marginal benefits help stabilize response length effectively.

%% file: sec/4_experiment.tex
\section{Experiments and Findings}
\label{sec:experiment}

\subsection{Experimental Setups}

Based on Qwen2.5VL-Instruct-7B~\cite{bai2025qwen25vl}, we conducted all RFT experiments on eight NVIDIA A100 80GB GPUs. The batch size is set to 1 per device, with gradient accumulation in 4 steps. For model generation, the temperature is set to 1, and the KL coefficient is fixed at 0.04. To investigate the impact of response length on model performance, we set the maximum response length at 1,024 tokens. We retain the built-in system prompt and incorporate format-related questions in the user prompt for GRPO training, where each sample generates 8 rollouts. The model is trained for 300 steps with a learning rate of 1e-6. For supervised fine-tuning, we adopt Qwen2.5VL-Instruct-7B and train it for 300 steps to ensure a fair comparison with RFT, using LLama-Factory framework~\cite{zheng2024llamafactory}.

\subsection{Quantitative Comparison over CV Tasks}
\input{Tables/main_tab}
We evaluate our model on three benchmarks: Banner Analysis, FGVC Aircraft ~\cite{fgvc}, and SAT~\cite{SAT}. Banner Analysis is a dataset that we collected for sentiment classification. It serves as a benchmark for detecting negative content in banners and slogans. The dataset comprises real-world images of various banners and contains 331 images for training, specifically curated for this task. As a binary classification benchmark, it takes a banner image as input and outputs either ``positive'' or ``negative'', indicating whether the banner contains positive or negative content. FGVC Aircraft~\cite{fgvc} is a fine-grained classification task that involves identifying airplanes across 100 distinct classes, totally 6k training images. SAT~\cite{SAT} is a spatial aptitude training dataset designed to challenge users with complex, dynamic spatial tasks that go beyond the static relationships found in traditional datasets. 15k data samples are used for training. These three benchmarks increase in difficulty, with a progressively higher demand for reasoning ability. 

In Table~\ref{tab:main}, we compare the effects of Supervised Fine-Tuning (SFT) and Reinforcement Fine-Tuning (RFT) on Qwen2.5VL-Instruct-7B~\cite{bai2025qwen25vl}. Both SFT and RFT models are trained on the corresponding datasets for 300 steps for fair comparison. We investigate the impact of explicitly guiding MLLMs to engage in reasoning during test by requiring three paradigms to generate reasoning process within \texttt{<think>} and \texttt{</think>} tokens.



The results demonstrate that incorporating \texttt{<think>} tokens in prompts leads to underperformance in both training-free and SFT models, revealing that extended intermediate reasoning steps may induce overthinking. In contrast, RFT models, while direct generating answer reduces accuracy, explicit reasoning steps substantially enhance performance. This paradigm shift implies that RFT training could enhance the model's structured reasoning capabilities.

These findings suggest that RFT is a more effective approach than SFT for these CV benchmarks. Additionally, explicit reasoning proves beneficial for complex benchmarks but offers limited advantages for simpler tasks.

\begin{figure}[t]
    \centering
    
    \begin{subfigure}{0.85\linewidth}
        \includegraphics[width=\linewidth]{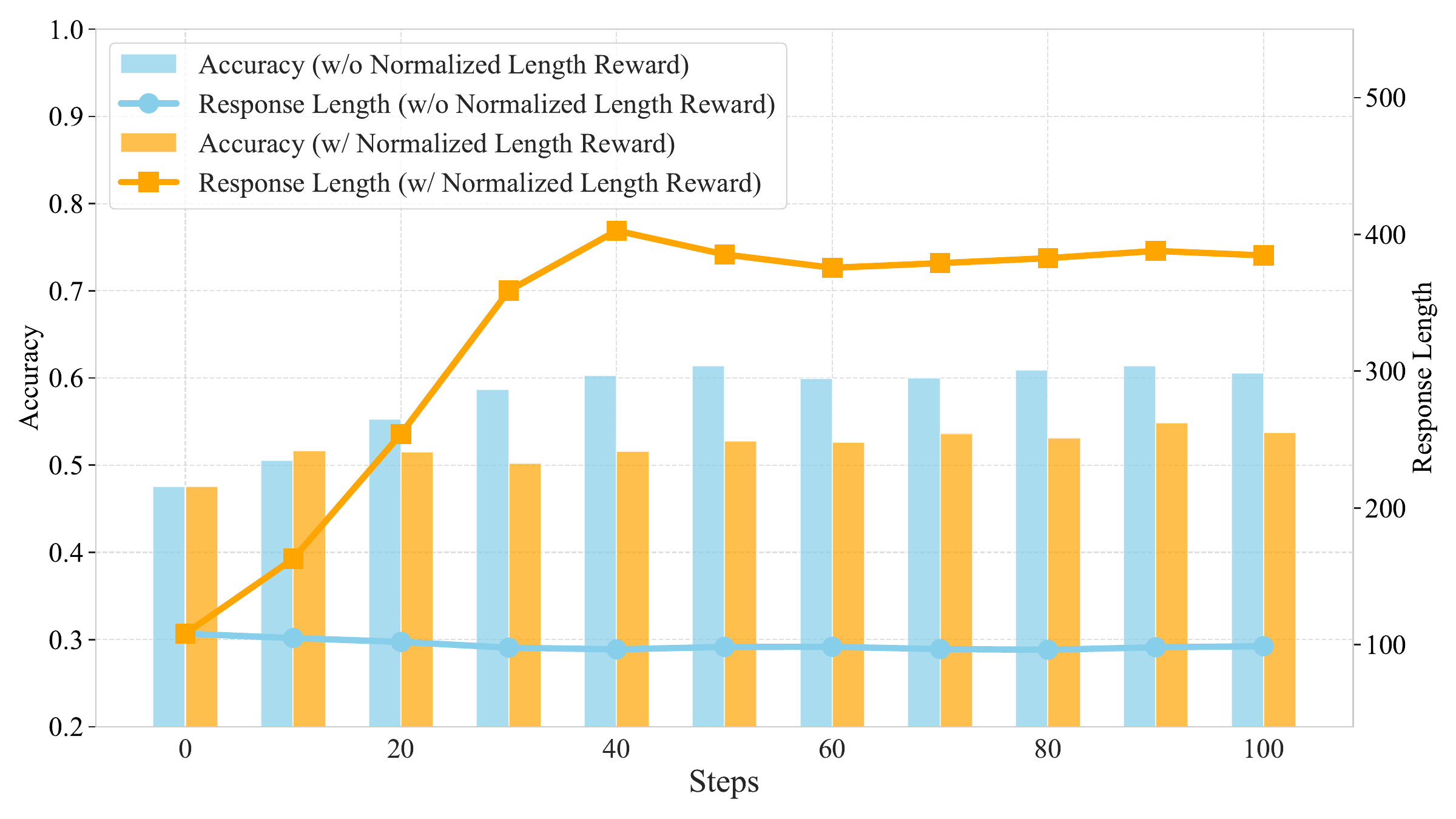}
        \caption{\textbf{FGVC Aircraft.}}
        \label{fig:fgvc}
    \end{subfigure}

    \begin{subfigure}{0.85\linewidth}
        \includegraphics[width=\linewidth]{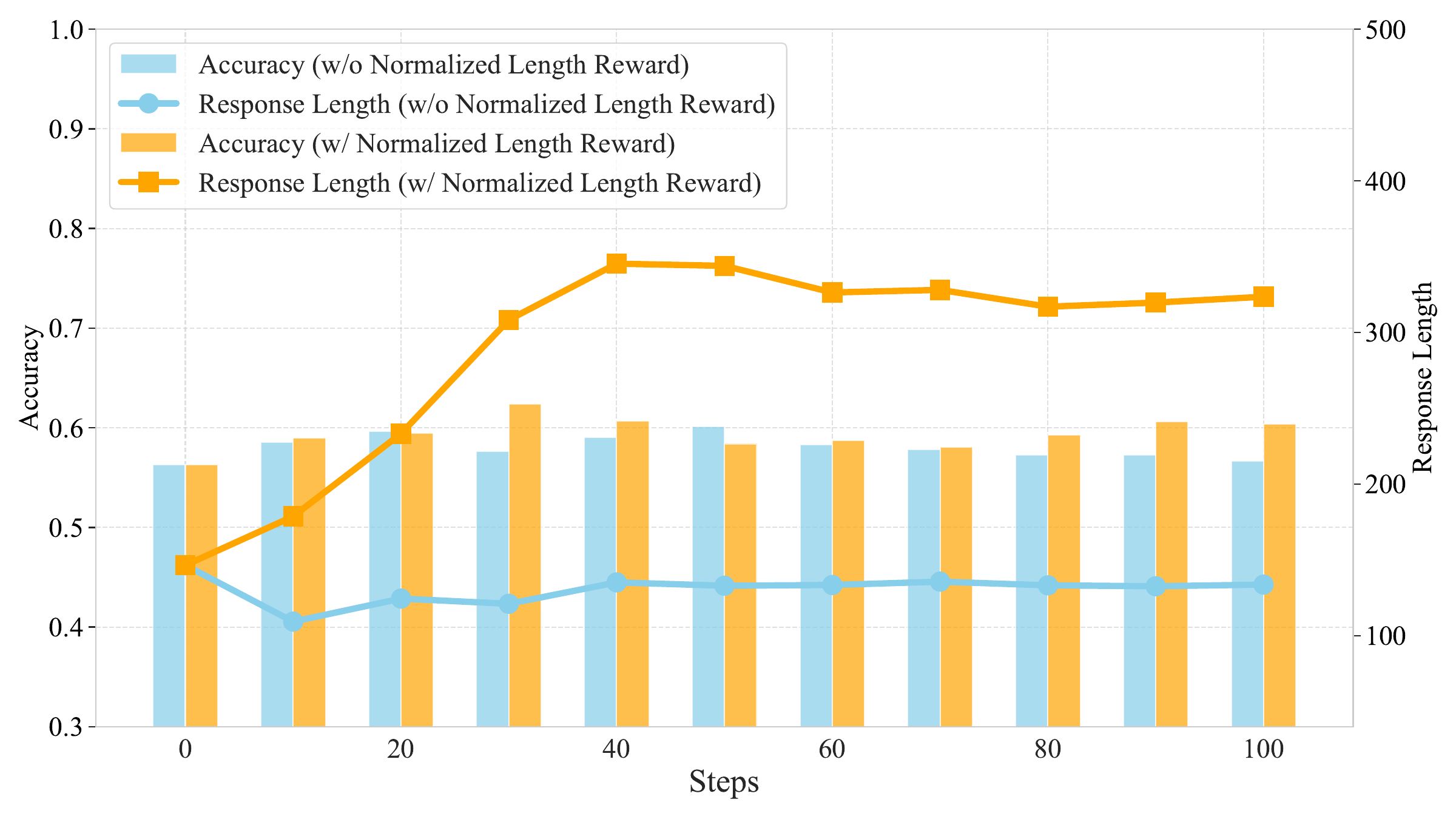}
        \caption{\textbf{SAT.}}
        \label{fig:sat}
    \end{subfigure}
    
    \caption{\textbf{Accuracy and Response Length over Steps.}}
    \vspace{-0.5cm}
    \label{fig:response-len}
\end{figure}

\subsection{Performance and Response Length}

In DeepSeek-R1~\cite{guo2025deepseekr1}, researchers identified an ``aha moment'' in Large Language Models (LLMs) from the model performance on text-only benchmarks: as response length increases, accuracy improves, suggesting that the model exhibits a capacity for self-reflection. However, it raises the question of whether computer vision (CV) task necessitates an extensive reasoning process. For example, most CV classification tasks primarily evaluate perceptual capabilities. And such tasks may not invariably benefit from prolonged explicit reasoning. Just as humans classify objects by recognizing salient features without elaborate reasoning, MLLMs may also gain little benefit from an extended reasoning process.

We incorporated a normalized length reward into the Reinforcement Fine-Tuning (RFT) process to investigate this hypothesis. In Figure~\ref{fig:response-len}, subfigures (a)  illustrate a relatively straightforward visual task, FGVC Airplane classification that only involves object classification based on the images provided. For such task, where visual perception is paramount, an increase in response length detrimentally affects performance compared to experiments conducted without the length reward. In contrast, within the SAT benchmark in Figure~\ref{fig:sat}, where reasoning is required after object recognition, longer responses correlate with improved performance.

Although the qualitative analysis presented in Figure~\ref{fig:vis} indicates that the application of the normalized length reward improves the reasoning process and logical coherence, the benchmark accuracy results reveal that the impact of increased response length on performance varies as a function of the difficulty and complexity of the task.

We found that encouraging Multimodal Large Language Models (MLLMs) to think longer during the Reinforcement Fine-Tuning (RFT) process does not lead to significant performance improvements. This suggests that current reasoning approaches are of limited help for traditional vision tasks. The reason may be that traditional Computer Vision (CV) tasks focus more on perceptual capabilities rather than complex reasoning processes.


%% file: Tables/main_tab.tex


\begin{table*}
\centering
\caption{\textbf{Performance Comparison Between SFT and RFT.} The three benchmarks listed has an increasing demand on reasoning ability. Banner Analysis is our constructed real-world benchmark for recognizing malicious content. }
\label{tab:main}
\begin{adjustbox}{width=0.9\linewidth}
\begin{tabular}{ccccccc} 
\toprule
\multicolumn{1}{l}{} & \multicolumn{6}{c}{Qwen2.5VL-Instruct-7B}                                              \\ 
\cmidrule(lr){2-3}\cmidrule(lr){4-5}\cmidrule(lr){6-7}
Benchmarks           & \multicolumn{2}{c}{Training Free} & \multicolumn{2}{c}{SFT} & \multicolumn{2}{c}{RFT}  \\ 
\cmidrule(lr){2-2}\cmidrule(lr){3-3}\cmidrule(lr){4-4}\cmidrule(lr){5-5}\cmidrule(lr){6-6}\cmidrule(lr){7-7}
\multicolumn{1}{l}{} & \texttt{w/o <think>} & \texttt{w/ <think>}              & \texttt{w/o <think>} & \texttt{w/ <think>}    & \texttt{w/o <think>} & \texttt{w/ <think>}     \\ 
\midrule
Banner-Analysis      & 80.72      & 80.11                
& 90.74     & 90.21      & 97.22     & \textbf{97.83}        \\
FGVC-Aircraft                 & 55.48     & 47.55                 & 75.85     & 75.84       & 63.57     & \textbf{83.32}        \\
SAT                  &  56.85    & 49.84                 & 58.24     & 59.62         & 61.01       & \textbf{62.23}          \\
\bottomrule
\end{tabular}
\end{adjustbox}
\end{table*}

%% file: sec/5_conclusion.tex
\section{Conclusions}
\label{sec:conclusion}

In this work, we perform experimental examination over the suitabilities and limitations of Reinforcement Fine-Tuning in visual tasks. Through quantitative comparisons, ablation studies and qualitative cases, we find that RFT generally works better than SFT for MLLMs over traditional CV tasks.
Furthermore, we find that though the thinking process from RFT is essential for MLLMs, encouraging them to think longer and producing more intermediate results is not always helpful, possibly due to the nature of these tasks for less ``reasoning'' than ``recognition''.
We hope this study will serve as a pilot study for the MLLM community where RL is only at the stage of early exploration.

%% file: main.bbl
\begin{thebibliography}{24}
\providecommand{\natexlab}[1]{#1}
\providecommand{\url}[1]{\texttt{#1}}
\expandafter\ifx\csname urlstyle\endcsname\relax
  \providecommand{\doi}[1]{doi: #1}\else
  \providecommand{\doi}{doi: \begingroup \urlstyle{rm}\Url}\fi

\bibitem[Bai et~al.(2025)Bai, Chen, Liu, Wang, Ge, Song, Dang, Wang, Wang, Tang, et~al.]{bai2025qwen25vl}
Shuai Bai, Keqin Chen, Xuejing Liu, Jialin Wang, Wenbin Ge, Sibo Song, Kai Dang, Peng Wang, Shijie Wang, Jun Tang, et~al.
\newblock Qwen2. 5-vl technical report.
\newblock \emph{arXiv preprint arXiv:2502.13923}, 2025.

\bibitem[Chen et~al.(2024)Chen, Wu, Wang, Su, Chen, Xing, Zhong, Zhang, Zhu, Lu, et~al.]{chen2024internvl}
Zhe Chen, Jiannan Wu, Wenhai Wang, Weijie Su, Guo Chen, Sen Xing, Muyan Zhong, Qinglong Zhang, Xizhou Zhu, Lewei Lu, et~al.
\newblock Internvl: Scaling up vision foundation models and aligning for generic visual-linguistic tasks.
\newblock In \emph{Proceedings of the IEEE/CVF conference on computer vision and pattern recognition}, pages 24185--24198, 2024.

\bibitem[Fu et~al.(2023)Fu, Chen, Shen, Qin, Zhang, Lin, Yang, Zheng, Li, Sun, et~al.]{fu2023mme}
Chaoyou Fu, Peixian Chen, Yunhang Shen, Yulei Qin, Mengdan Zhang, Xu Lin, Jinrui Yang, Xiawu Zheng, Ke Li, Xing Sun, et~al.
\newblock Mme: A comprehensive evaluation benchmark for multimodal large language models.
\newblock \emph{arXiv preprint arXiv:2306.13394}, 2023.

\bibitem[Guo et~al.(2025)Guo, Yang, Zhang, Song, Zhang, Xu, Zhu, Ma, Wang, Bi, et~al.]{guo2025deepseekr1}
Daya Guo, Dejian Yang, Haowei Zhang, Junxiao Song, Ruoyu Zhang, Runxin Xu, Qihao Zhu, Shirong Ma, Peiyi Wang, Xiao Bi, et~al.
\newblock Deepseek-r1: Incentivizing reasoning capability in llms via reinforcement learning.
\newblock \emph{arXiv preprint arXiv:2501.12948}, 2025.

\bibitem[Huang et~al.(2025)Huang, Jia, Zhai, Cao, Ye, Zhao, Hu, and Lin]{huang2025visionr1}
Wenxuan Huang, Bohan Jia, Zijie Zhai, Shaosheng Cao, Zheyu Ye, Fei Zhao, Yao Hu, and Shaohui Lin.
\newblock Vision-r1: Incentivizing reasoning capability in multimodal large language models.
\newblock \emph{arXiv preprint arXiv:2503.06749}, 2025.

\bibitem[Hurst et~al.(2024)Hurst, Lerer, Goucher, Perelman, Ramesh, Clark, Ostrow, Welihinda, Hayes, Radford, et~al.]{hurst2024gpt}
Aaron Hurst, Adam Lerer, Adam~P Goucher, Adam Perelman, Aditya Ramesh, Aidan Clark, AJ Ostrow, Akila Welihinda, Alan Hayes, Alec Radford, et~al.
\newblock Gpt-4o system card.
\newblock \emph{arXiv preprint arXiv:2410.21276}, 2024.

\bibitem[Lai et~al.(2024)Lai, Tian, Chen, Li, Yuan, Liu, and Jia]{lai2024lisa}
Xin Lai, Zhuotao Tian, Yukang Chen, Yanwei Li, Yuhui Yuan, Shu Liu, and Jiaya Jia.
\newblock Lisa: Reasoning segmentation via large language model.
\newblock In \emph{Proceedings of the IEEE/CVF Conference on Computer Vision and Pattern Recognition}, pages 9579--9589, 2024.

\bibitem[Li et~al.(2024)Li, Ge, Ge, Wang, Wang, Zhang, and Shan]{li2024seed}
Bohao Li, Yuying Ge, Yixiao Ge, Guangzhi Wang, Rui Wang, Ruimao Zhang, and Ying Shan.
\newblock Seed-bench: Benchmarking multimodal large language models.
\newblock In \emph{Proceedings of the IEEE/CVF Conference on Computer Vision and Pattern Recognition}, pages 13299--13308, 2024.

\bibitem[Liu et~al.(2023)Liu, Li, Wu, and Lee]{liu2023llava}
Haotian Liu, Chunyuan Li, Qingyang Wu, and Yong~Jae Lee.
\newblock Visual instruction tuning.
\newblock \emph{Advances in neural information processing systems}, 36:\penalty0 34892--34916, 2023.

\bibitem[Liu et~al.(2024{\natexlab{a}})Liu, Chen, Cai, Jiang, Hu, Yao, Wang, and Xie]{liu2024lamra}
Yikun Liu, Pingan Chen, Jiayin Cai, Xiaolong Jiang, Yao Hu, Jiangchao Yao, Yanfeng Wang, and Weidi Xie.
\newblock Lamra: Large multimodal model as your advanced retrieval assistant.
\newblock \emph{arXiv preprint arXiv:2412.01720}, 2024{\natexlab{a}}.

\bibitem[Liu et~al.(2024{\natexlab{b}})Liu, Duan, Zhang, Li, Zhang, Zhao, Yuan, Wang, He, Liu, et~al.]{liu2024mmbench}
Yuan Liu, Haodong Duan, Yuanhan Zhang, Bo Li, Songyang Zhang, Wangbo Zhao, Yike Yuan, Jiaqi Wang, Conghui He, Ziwei Liu, et~al.
\newblock Mmbench: Is your multi-modal model an all-around player?
\newblock In \emph{European conference on computer vision}, pages 216--233. Springer, 2024{\natexlab{b}}.

\bibitem[Liu et~al.(2025)Liu, Sun, Zang, Dong, Cao, Duan, Lin, and Wang]{liu2025visualrft}
Ziyu Liu, Zeyi Sun, Yuhang Zang, Xiaoyi Dong, Yuhang Cao, Haodong Duan, Dahua Lin, and Jiaqi Wang.
\newblock Visual-rft: Visual reinforcement fine-tuning.
\newblock \emph{arXiv preprint arXiv:2503.01785}, 2025.

\bibitem[Maji et~al.(2013)Maji, Rahtu, Kannala, Blaschko, and Vedaldi]{fgvc}
Subhransu Maji, Esa Rahtu, Juho Kannala, Matthew Blaschko, and Andrea Vedaldi.
\newblock Fine-grained visual classification of aircraft.
\newblock \emph{arXiv preprint arXiv:1306.5151}, 2013.

\bibitem[Meng et~al.(2025)Meng, Du, Liu, Zhou, Lu, Fu, Shi, Wang, He, Zhang, et~al.]{meng2025mmeureka}
Fanqing Meng, Lingxiao Du, Zongkai Liu, Zhixiang Zhou, Quanfeng Lu, Daocheng Fu, Botian Shi, Wenhai Wang, Junjun He, Kaipeng Zhang, et~al.
\newblock Mm-eureka: Exploring visual aha moment with rule-based large-scale reinforcement learning.
\newblock \emph{arXiv preprint arXiv:2503.07365}, 2025.

\bibitem[Mitra et~al.(2024)Mitra, Huang, Chai, Lin, Arbelle, Feris, Karlinsky, Darrell, Ramanan, and Herzig]{mitra2024sparse}
Chancharik Mitra, Brandon Huang, Tianning Chai, Zhiqiu Lin, Assaf Arbelle, Rogerio Feris, Leonid Karlinsky, Trevor Darrell, Deva Ramanan, and Roei Herzig.
\newblock Sparse attention vectors: Generative multimodal model features are discriminative vision-language classifiers.
\newblock \emph{arXiv preprint arXiv:2412.00142}, 2024.

\bibitem[Ouali et~al.(2024)Ouali, Bulat, Xenos, Zaganidis, Metaxas, Martinez, and Tzimiropoulos]{ouali2024discriminative}
Yassine Ouali, Adrian Bulat, Alexandros Xenos, Anestis Zaganidis, Ioannis~Maniadis Metaxas, Brais Martinez, and Georgios Tzimiropoulos.
\newblock Discriminative fine-tuning of lvlms.
\newblock \emph{arXiv preprint arXiv:2412.04378}, 2024.

\bibitem[Ouyang et~al.(2022)Ouyang, Wu, Jiang, Almeida, Wainwright, Mishkin, Zhang, Agarwal, Slama, Ray, et~al.]{ouyang2022rlhf}
Long Ouyang, Jeffrey Wu, Xu Jiang, Diogo Almeida, Carroll Wainwright, Pamela Mishkin, Chong Zhang, Sandhini Agarwal, Katarina Slama, Alex Ray, et~al.
\newblock Training language models to follow instructions with human feedback.
\newblock \emph{Advances in neural information processing systems}, 35:\penalty0 27730--27744, 2022.

\bibitem[Pi et~al.(2023)Pi, Gao, Diao, Pan, Dong, Zhang, Yao, Han, Xu, Kong, et~al.]{pi2023detgpt}
Renjie Pi, Jiahui Gao, Shizhe Diao, Rui Pan, Hanze Dong, Jipeng Zhang, Lewei Yao, Jianhua Han, Hang Xu, Lingpeng Kong, et~al.
\newblock Detgpt: Detect what you need via reasoning.
\newblock \emph{arXiv preprint arXiv:2305.14167}, 2023.

\bibitem[Ray et~al.(2024)Ray, Duan, Tan, Bashkirova, Hendrix, Ehsani, Kembhavi, Plummer, Krishna, Zeng, et~al.]{SAT}
Arijit Ray, Jiafei Duan, Reuben Tan, Dina Bashkirova, Rose Hendrix, Kiana Ehsani, Aniruddha Kembhavi, Bryan~A Plummer, Ranjay Krishna, Kuo-Hao Zeng, et~al.
\newblock Sat: Spatial aptitude training for multimodal language models.
\newblock \emph{arXiv preprint arXiv:2412.07755}, 2024.

\bibitem[Schulman et~al.(2017)Schulman, Wolski, Dhariwal, Radford, and Klimov]{schulman2017ppo}
John Schulman, Filip Wolski, Prafulla Dhariwal, Alec Radford, and Oleg Klimov.
\newblock Proximal policy optimization algorithms.
\newblock \emph{arXiv preprint arXiv:1707.06347}, 2017.

\bibitem[Shao et~al.(2024)Shao, Wang, Zhu, Xu, Song, Bi, Zhang, Zhang, Li, Wu, et~al.]{shao2024deepseekmath}
Zhihong Shao, Peiyi Wang, Qihao Zhu, Runxin Xu, Junxiao Song, Xiao Bi, Haowei Zhang, Mingchuan Zhang, YK Li, Y Wu, et~al.
\newblock Deepseekmath: Pushing the limits of mathematical reasoning in open language models.
\newblock \emph{arXiv preprint arXiv:2402.03300}, 2024.

\bibitem[Team et~al.(2025)Team, Du, Gao, Xing, Jiang, Chen, Li, Xiao, Du, Liao, et~al.]{team2025kimik15}
Kimi Team, Angang Du, Bofei Gao, Bowei Xing, Changjiu Jiang, Cheng Chen, Cheng Li, Chenjun Xiao, Chenzhuang Du, Chonghua Liao, et~al.
\newblock Kimi k1. 5: Scaling reinforcement learning with llms.
\newblock \emph{arXiv preprint arXiv:2501.12599}, 2025.

\bibitem[Zheng et~al.(2024)Zheng, Zhang, Zhang, Ye, Luo, Feng, and Ma]{zheng2024llamafactory}
Yaowei Zheng, Richong Zhang, Junhao Zhang, Yanhan Ye, Zheyan Luo, Zhangchi Feng, and Yongqiang Ma.
\newblock Llamafactory: Unified efficient fine-tuning of 100+ language models.
\newblock In \emph{Proceedings of the 62nd Annual Meeting of the Association for Computational Linguistics (Volume 3: System Demonstrations)}, Bangkok, Thailand, 2024. Association for Computational Linguistics.

\bibitem[Zhou et~al.(2025)Zhou, Li, Wang, Cheng, Zhou, and Hsieh]{zhou2025r1visualthinker}
Hengguang Zhou, Xirui Li, Ruochen Wang, Minhao Cheng, Tianyi Zhou, and Cho-Jui Hsieh.
\newblock R1-zero's" aha moment" in visual reasoning on a 2b non-sft model.
\newblock \emph{arXiv preprint arXiv:2503.05132}, 2025.

\end{thebibliography}
